\documentclass[twoside]{article}
\usepackage{cite}
\usepackage{caption}
\usepackage{PRIMEarxiv}
\usepackage{amsmath}
\usepackage[utf8]{inputenc} 
\usepackage[T1]{fontenc}    
\usepackage{hyperref}       
\usepackage{url}            
\usepackage{booktabs}       
\usepackage{amsfonts}       
\usepackage{nicefrac}       
\usepackage{microtype}      
\usepackage{lipsum}
\usepackage{fancyhdr}       
\usepackage{graphicx}       
\graphicspath{{media/}}     
\usepackage{threeparttable}
\title{A Novel Dual-Stream Framework for dMRI Tractography Streamline Classification with Joint dMRI and fMRI Data
}

\author{
  Haotian Yan$^{1,*}$, 
  Bocheng Guo$^{1,*}$ \\
  $^{1}$University of Electronic Science and Technology of China, Chengdu, China\\
   \And
  Jianzhong He$^{2}$ \\
  $^{2}$Zhejiang University of Technology, Hangzhou, China \\
  \And
  Nir A. Sochen$^{3}$ \\
  $^{3}$University of Tel Aviv, Israel \\
  \And
  Ofer Pasternak$^{4}$, 
  Lauren J O’Donnell$^{4}$ \\
  $^{4}$Harvard Medical School, Boston, USA \\
  \And
  Fan Zhang$^{\#}$ \\
  $^{1}$University of Electronic Science and Technology of China, Chengdu, China\\
}

\begin{document}
\maketitle
{
  \renewcommand\thefootnote{} 
  \footnotetext{*Haotian Yan and Bocheng Guo are co-first-authors}
  \footnotetext{\#Corresponding to author Fan Zhang (fan.zhang@uestc.edu.cn)}
} 
\begin{abstract}
Streamline classification is essential to identify anatomically meaningful white matter tracts from diffusion MRI (dMRI) tractography. However, current streamline classification methods rely primarily on the geometric features of the streamline trajectory, failing to distinguish between functionally distinct fiber tracts with similar pathways. To address this, we introduce a novel dual-stream streamline classification framework that jointly analyzes dMRI and functional MRI (fMRI) data to enhance the functional coherence of tract parcellation. We design a novel network that performs streamline classification using a pretrained backbone model for full streamline trajectories, while augmenting with an auxiliary network that processes fMRI signals from fiber endpoint regions. We demonstrate our method by parcellating the corticospinal tract (CST) into its four somatotopic subdivisions. Experimental results from ablation studies and comparisons with state-of-the-art methods demonstrate our approach's superior performance.
\end{abstract}


\section{Introduction}
\label{sec:intro}

Diffusion MRI (dMRI) tractography \cite{vivo} is currently the only method that enables non-invasive mapping of the brain's white matter (WM) connections \textit{in vivo.} It is an important neuroimaging tool to quantify the brain's wiring diagram in health and disease \cite{quantitative}. A key computational task in tractography is streamline classification, where the reconstructed streamlines are assigned with anatomical or functional labels to segment specific fiber tracts. This is crucial for identifying anatomically meaningful fiber tracts and enabling robust analyses across populations \cite{anato,develop,ciccarelli2008diffusion}. However, existing streamline classification approaches often focus on separating large fiber bundles but face difficulties in the more nuanced task of differentiating functionally distinct subdivisions residing within the same tract.

A prime example is the corticospinal tract (CST), the brain's primary pathway for voluntary motor control \cite{cst}. The CST possesses a precise somatotopic organization, where distinct fiber populations originating from the motor cortex correspond to the control of the face, hand, leg, and trunk \cite{reconstruct}. Accurately classifying CST subdivisions is critical for both neuroscience research and clinical practice \cite{2007Quantitative,genetic}. Traditional methods, which rely on expert-drawn or atlas-derived regions of interest (ROIs), are often inefficient. For instance, one widely used technique for CST parcellation \cite{reconstruct} leverages FreeSurfer's cortical surface parcellation derived from T1-weighted images, a process that can take several hours to complete.

Deep learning offers an effective and efficient alternative for streamline classification in large-scale datasets. However, existing methods rely primarily on dMRI-derived streamline geometric features. For example, methods such as SupWMA \cite{supwma} and DCNN \cite{dcnn} analyze the spatial coordinates along the trajectory of individual streamlines to learn their shape representation, and TractCloud \cite{tractcloud} leverages the relative geometry within a local neighborhood by comparing a target streamline to its nearby streamlines. While powerful, these geometry-centric methods suffer from a common limitation: geometric similarity does not always imply functional specificity. Adjacent fiber populations within the CST with different functions (e.g., hand vs trunk subdivisions) may have highly similar spatial trajectories, making them difficult to distinguish using geometric information alone \cite{reconstruct}. Consequently, there is a compelling need to integrate complementary data modalities, such as brain functional information, to overcome the inherent ambiguities of purely geometric approaches.

Functional MRI (fMRI) offers a promising solution by capturing the functional dynamics of cortical areas near fiber tract endpoints via blood-oxygen-level-dependent (BOLD) signals. It provides complementary functional information that is absent in dMRI. However, fusing these two modalities for streamline classification is a challenging task. First, fMRI signals bring rich information with the advantage of high temporal resolution, but also comes with the challenge of high-dimensional and noise-laden data. Second, fMRI functional features differ greatly from dMRI structural features, and naive fusion can easily compromise the stability of the robust geometric information. Given these challenges, a rational and effective scheme for how to fuse these two modalities for classification has not yet been established.

In light of the above, we propose a novel dual-stream framework to robustly integrate fMRI data into dMRI streamline classification for tract parcellation. The overall idea is to perform streamline classification using a pretrained backbone model for full streamline trajectories, while augmenting with an auxiliary network that processes fMRI signals from fiber endpoint regions. There are two key innovations. First, to solve the fusion instability, we design a novel dual-stream architecture with a pre-trained geometric backbone with parameters frozen and an auxiliary functional pathway to process fMRI signals. This provides a stable geometric-based decision, and the auxiliary stream is trained separately to produce a corrective functional-based decision. These two decisions are fused at the class decision-level to ensure the robust geometric features are not corrupted by the functional signals during backpropagation. Second, to effectively extract features from the high-dimensional fMRI signal without information loss, our auxiliary pathway simultaneously processes the spatial location and fMRI signals at the endpoint regions. This preserves temporal dynamics and is tolerant to the streamline's start/end point order. This functional feature is then mutually constrained by the 3D endpoint location features (via an 'Endpoint Feature Integrator') to yield a highly efficient representation. Experimental results from ablation studies and comparisons with state-of-the-art methods demonstrate our approach's superior performance.

\begin{figure}[htbp]
    \centering
    
    \includegraphics[width=0.6\linewidth]{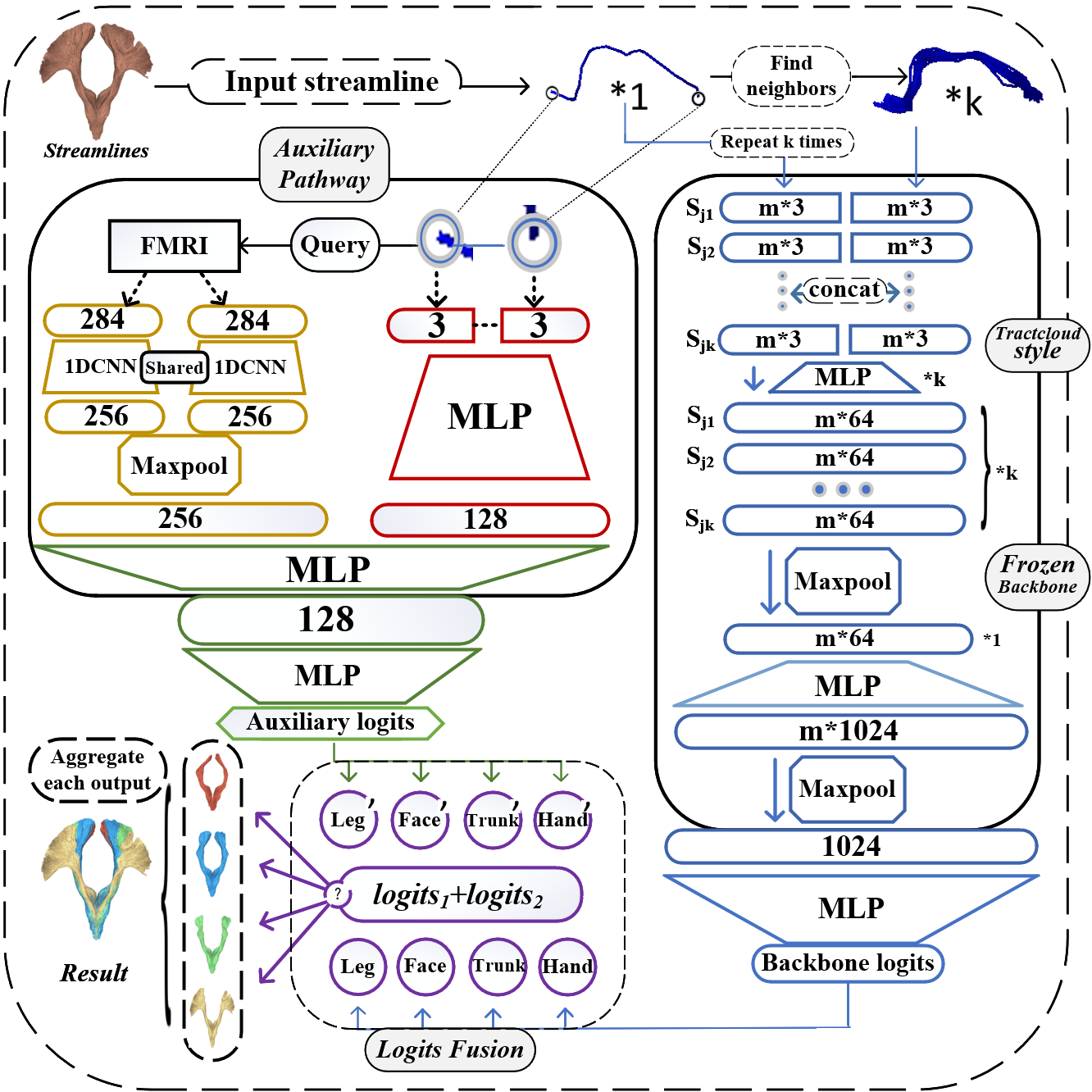}
    
    \caption{Overview of our model pipeline}
    
    \label{fig:example}
\end{figure}
\section{Methods}
\label{sec:method}

Our method is a multimodal fusion framework featuring a novel dual-stream architecture designed for streamline classification, with a geometric backbone to process streamline trajectory information and an auxiliary functional pathway to process fMRI BOLD signals. Figure 1 gives a method overview. The process begins by initializing the geometric backbone with a pre-training phase, where it learns to classify streamlines using only their full trajectory data. For the main network training, this backbone is frozen. Its output logits for the target classes are then integrated with a trainable auxiliary pathway that leverages streamline endpoint locations and fMRI information. Finally, the outputs of the geometric backbone and the auxiliary functional pathway are fused to compute the final classification prediction.\\\\
\textbf{2.1. Geometric Backbone}\\\\
The geometric backbone module utilizes the full trajectory geometry of tractography streamlines to perform an initial classification. This module is built upon our recently proposed TractCloud network \cite{tractcloud}, which can process the spatial coordinates of an individual streamline and further enrich this representation with local neighborhood information by comparing the target streamline to its proximal streamlines. Specifically, the input consists of the target streamline (resampled to 25 points) and its 20 nearest neighbors. To capture the local neighborhood context, the network computes pairwise feature\textbf{s} by concatenating the repeated coordinates of the target streamline with each of its 20 neighbors. These paired features are processed by a shared MLP to generate a feature map. This map is then aggregated via max-pooling (across the 20 neighbors) to create a robust local context representation, which is finally fed into a PointNet-style module \cite{pointnet} to produce a 1024-dimensional global geometric feature vector. Finally, a MLP classifier is used to produce the classification logits, \textit{logits}\textit{\textsubscript{backbone}}, which is used in dual-stream dMRI-fMRI fusion (Section 2.3). This geometric backbone’s parameters are frozen post-training.\\
\\
\textbf{2.2. Auxiliary Function Pathway}\\\\
The auxiliary pathway is incorporated to leverage fMRI features, facilitating multimodal analysis that jointly utilizes dMRI and fMRI data for robust streamline classification. Unlike the geometric backbone that is frozen to preserve the geometric prior, the auxiliary pathway is fully trainable, enabling a functional refinement of the geometry-based classification results. In our study, we focus on the streamline endpoint regions, leveraging the principle that functional specificity arises from this anatomical interface, particularly at the cortical surface. The network consists of a dual-encoder structure for feature extraction, followed by a fusion module to compute the final functional feature, as follows:

\textbf{1. Endpoint Geometric Encoder}: An MLP layer maps the endpoint coordinate vector—a 6D vector composed of the RAS coordinates of the two endpoints—into a 128-dimensional embedding to capture the spatial distribution of the streamline.

\textbf{2. Endpoint Functional Encoder}: A 1D-CNN to transform the fMRI signals from both endpoints into a 256-dimensional feature to capture the salient temporal patterns and activation profiles within the signals. To distinguish between the two endpoints, the fMRI signals from each are processed by separate 1D-CNNs with shared weights.

\textbf{3. Endpoint Feature Integrator}: An MLP to compute the final 128-dimensional auxiliary feature vector by integrating geometric and functional endpoint features. This approach uses spatial context to disambiguate and refine the functional information provided by the fMRI data.\\
\\
\textbf{2.3. Dual-Stream dMRI-fMRI Fusion}\\\\
The output features from the geometric backbone and the auxiliary function pathway are then fused to produce the final class prediction. Instead of a direct concatenation, each of the output features is processed by an additional MLP to generate 4-D logits (i.e., \textit{logits}\textit{\textsubscript{backbone}} and \textit{logits}\textit{\textsubscript{auxiliary}}), where each corresponds to one target class. Then, the output  fusion is computed by element-wise addition of the two logits, i.e.,
\begin{equation}
\text{logits}_{\text{final}} = \text{logits}_{\text{backbone}} + \text{logits}_{\text{auxiliary}}
\end{equation}
from which the final class is predicted by selecting the index with the maximum value. This fusion strategy contributes to our framework's stability: because the backbone network is frozen, gradients only flow back to update the auxiliary pathway. This architecture effectively prevents the noisy fMRI signal from corrupting the robust, pre-learned geometric knowledge of the baseline, while still allowing the functional data to provide a corrective refinement. \\
\\
\textbf{2.4. Losses and Implementations}\\\\
Both the geometric backbone's pretraining and the overall network's training use the same loss. To handle the imbalanced distribution of streamlines across the four classes, we employ a weighted cross-entropy loss function, with the weights \\
\begin{equation}
w_c = \frac{N}{c \times n_c}
\end{equation}
where \textit{N} is the total number of streamlines, \textit{c} is the number of classes (4 in our case), and \textit{n}\textit{\textsubscript{c}} is the number of streamline in the class. 

We implement our framework using PyTorch 2.5.1+cu121 and the computation is performed on a Nvidia 3090 GPU server. The backbone model is trained for 30 epochs with a batch size of 512. The Adam optimizer is used with a learning rate of 1e-4 and a Cosine annealing scheduler. During overall network training, the backbone model is loaded and its weights are frozen. The network is then trained for 20 epochs, using the same optimization settings as the backbone network (Adam, LR=1e-4, Cosine annealing, bath size 512). 

\section{EXPERIMENTS AND RESULTS}
\label{sec:pagestyle}

\textbf{3.1. Data and Preprocessing}\\\\
In this paper, we use the pre-preprocessed dMRI and motor task fMRI data from the unrelated 100 subjects in the Human Connectome Project Young Adult (HCP-YA) \cite{minimal}. For each subject, the dMRI data was acquired with a spatial resolution of 1.25×1.25×1.25 mm\textsuperscript{3}, TR/TE=5520/89.5 ms, and 270 directions across b-values of 1000/2000/3000 s/mm²; the tfMRI data was acquired with a spatial resolution of 2×2×2 mm\textsuperscript{3}, TR/TE=720/33.1 ms. 

dMRI Processing: We use TractSeg \cite{tractseg} to compute the whole CST with 200,000 streamlines per subject. To generate ground-truth labels for each streamline, we apply ROI-based selection to divide the CST into four subdivisions (i.e., leg, trunk, face and hand regions) using a cortical parcellation atlas via the FreeSurfer software, as described in \cite{reconstruct}. All streamlines are resampled to 25 points, and 20 nearest neighbors are identified for the TractCloud backbone input.

fMRI Preprocessing: We use the tfMRI data that has been preprocessed in HCP-YA. To further reduce the noise in the fMRI signals, we design a mask-guided joint spatial-temporal tfMRI denoising pipeline. First, a cortical mask is computed based on the endpoint regions of the CST. Then, tfMRI signals within the mask  undergo our composite filtering process: (1) a 6mm FWHM Gaussian spatial smooth; (2) a 0.01Hz high-pass temporal filter to remove signal drift; and (3) a 3×3×3 (6mm) boxcar (regional average) spatial filter. The processed tfMRI signals are then mapped to the streamlines that pass through the same voxels.\\
\\
\textbf{3.2. Experimental Design and Results}\\\\
We perform two experiments, including an ablation study to assess the proposed network architecture and a comparison with SOTA methods. All experiments are conducted using a 5-fold cross-validation scheme. Weighted F1-score is used as the quantitative evaluation metric.
\\
\\
\textbf{3.2.1. Ablation Comparison}\\\\
We perform an ablation study to evaluate our proposed components (Table 1). To demonstrate the generalizability of our fusion module, these ablations are performed on two distinct geometric backbones: PointNet (a fundamental network using only streamline coordinates) and Tractcloud (our main backbone, which also uses neighbor information). We compare the following configurations:\\
(1) None (Baseline): The geometry-only backbone (PointNet or Tractcloud) trained without any auxiliary pathway.\\
(2) + Endpoint only w. Logits Fusion: The Baseline fused via Logits Fusion with an auxiliary pathway processing only 3D endpoint coordinates.\\
(3) + fMRI only w. Logits Fusion: The Baseline fused via Logits Fusion with an auxiliary pathway processing only the fMRI signal.\\
(4) + fMRI+endpoint w. Conc Fusion: The Baseline fused at the feature-level (concatenation) with the full auxiliary pathway (fMRI+endpoint).\\
(5) + fMRI+endpoint w. Logits Fusion (Proposed): Our full method, fusing the Baseline and the full auxiliary pathway at the logit-level.

As shown in Table 1, the Baseline models (1) set the initial performance. Adding Endpoint only (2) or fMRI only (3) yields no significant improvement. The Conc Fusion strategy (4) results in performance degradation, demonstrating that naive feature-level fusion is detrimental. Our Proposed Logits Fusion (5) is the only method to achieve a consistent and statistically significant improvement ($p < 0.01$) on both backbones, confirming that both the Logits Fusion architecture and the (fMRI+Endpoint) geometric constraint are essential for success. This quantitative superiority is also supported by the visual comparison in Fig. 2, which shows our proposed method (blue) is qualitatively more accurate and functionally coherent with the fMRI background signal than the geometry-only baseline (yellow). \\
\\
  \begin{threeparttable}
    \centering
    \resizebox{0.7\linewidth}{!}{%
      \begin{tabular}{llcc}
        \toprule
        \textbf{Backbone Model} & \textbf{Auxiliary Pathway / Fusion Strategy} & \textbf{F1 (Mean $\pm$ Std)} \\
        \midrule

        PointNet            & None (Baseline)                     & 0.8705 ($\pm$ 0.0061)  \\
        PointNet            & + Endpoint only w. Logits Fusion    & 0.8704 ($\pm$ 0.0075)  \\
        PointNet            & + fMRI only w. Logits Fusion        & 0.8741 ($\pm$ 0.0104)  \\
        PointNet            & + fMRI+endpoint w. Conc Fusion      & 0.8676 ($\pm$ 0.0070)  \\
        PointNet            & \textbf{+ fMRI+endpoint w. Logits Fusion} & \textbf{0.8858 ($\pm$ 0.0077)}  \\
        \midrule
        Tractcloud          & None (Baseline)                     & 0.8973 ($\pm$ 0.0088) &   \\
        Tractcloud          & + Endpoint only w. Logits Fusion    & 0.8974 ($\pm$ 0.0097)  \\
        Tractcloud          & + fMRI only w.  Logits Fusion       & 0.8980 ($\pm$ 0.0090)  \\
        Tractcloud          & + fMRI+endpoint w. Conc Fusion      & 0.8942 ($\pm$ 0.0085)  \\
        
        Tractcloud          & \textbf{+ fMRI+endpoint w. Logits Fusion }    & \textbf{0.9015 ($\pm$ 0.0098)}  \\

        \bottomrule
      \end{tabular}
    }
\begin{flushleft} \small
\caption{Ablation study. }
\end{flushleft}
\end{threeparttable}

\begin{center} 
  \includegraphics[width=0.6\linewidth]{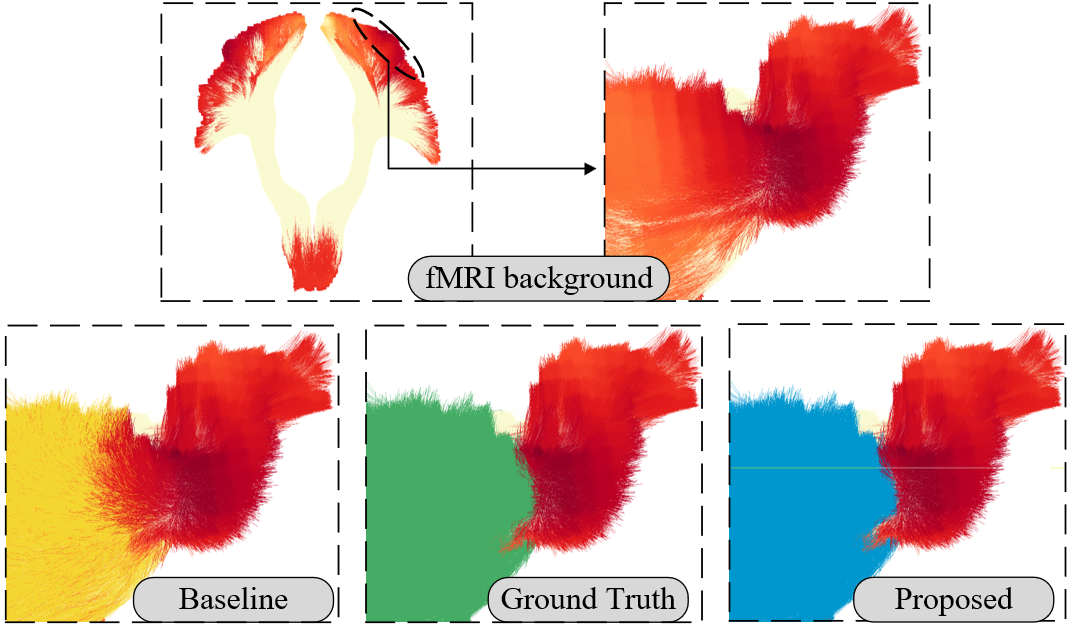}
  
  \captionof{figure}{The top row is the CST region colored by PCA-reduced fMRI endpoint signals, and the bottom row is the comparison of the prediction results for 'trunk' class from the geometry-only model and the proposed model against its ground truth region.   }
  \label{Fig1:visable results}

\end{center}
\textbf{3.2.3. SOTA Comparison\\\\}
We finally evaluate our full framework against several state-of-the-art methods for streamline classification, including: DCNN \cite{dcnn} and DeepWMA \cite{deepwma} that perform streamline classification using streamline geometry, and DMVFC \cite{dmvfc} that performs joint fMRI-dMRI-based streamline parcellation (DMVFC was designed for unsupervised streamline clustering; here we modify it for supervised streamline classification leveraging the fusion strategy). As shown in Table 2, DeepWMA achieves an F1-score of 0.8559, DCNN reaches 0.8785, and DMVFC obtains 0.8705. In contrast, our proposed method yields the highest F1-score of 0.9015, outperforming all these methods. This consistent performance gain underscores the superiority of our design—by explicitly integrating multi-modal information through the proposed fusion strategy, our framework effectively captures the complementary insights from geometric and functional data.\\

  \begin{threeparttable}

    \centering 
    \resizebox{0.7\linewidth}{!}{%
      \begin{tabular}{llcc}
        \toprule
        \textbf{Model} & \textbf{ Fusion Strategy} & \textbf{F1 (Mean $\pm$ Std)} \\
        \midrule

                DeepWMA                 & None                     & 0.8559 ($\pm$ 0.0089)  \\
        DCNN                  & None                      & 0.8785 ($\pm$ 0.0039)  \\
        DMVFC (Variant)
        &Multi-view Collaborative learning
        &0.8705 ($\pm$ 0.0086) \\
        \textbf{Proposed} &\textbf{Logits Fusion} &\textbf{0.9015 ($\pm$ 0.0098)}  \\
        \bottomrule
      \end{tabular}%
    } 
\begin{flushleft} \small 
\caption{Performance comparison with SOTA. }
\end{flushleft}
  \end{threeparttable}
\section{ Conclusion}
\label{sec:print1}

In this paper, our proposed method tackles the problem of combined dMRI-fMRI analysis for accurate streamline classification. We propose a novel dual-stream fusion method that can effectively use the full streamline trajectory information and the fMRI signals at the streamline endpoint regions. We demonstrate successful parcellation of the CST into four distinct subdivisions that reveals its somatotopic organization.

\section{COMPLIANCE WITH ETHICAL STANDARDS}
\label{sec:print}

This study was conducted retrospectively using public HCP imaging data. No ethical approval was required.
\section*{Acknowledgments}
This work is in part supported by the National Key RD Program of China (No. 2023YFE0118600), the National Natural Science Foundation of China (No. 62371107).

\bibliographystyle{unsrt}  
\bibliography{references}

\end{document}